\documentclass{article}

\usepackage{microtype}
\usepackage{graphicx}
\usepackage{subcaption}
\usepackage{booktabs} 

\usepackage{hyperref}

\usepackage[preprint]{icml2026}

\usepackage{hyperref}
\usepackage{url}
\usepackage{microtype}
\usepackage{graphicx}
\usepackage{booktabs}
\usepackage{float}
\usepackage{amsmath}
\usepackage{amssymb}
\usepackage{mathtools}
\usepackage{amsthm}
\usepackage{bm}
\usepackage{algorithm}
\usepackage{algorithmic}
\usepackage{multirow}
\usepackage{graphicx}
\usepackage{caption}
\usepackage{subcaption}
\usepackage{wrapfig}
\usepackage{tikz}
\usepackage{fancybox}
\usepackage{xcolor}
\usepackage{colortbl}
\usepackage{empheq}
\usepackage{multirow}
\usepackage{tabularx}
\usepackage[shortlabels,inline]{enumitem}
\usepackage{tcolorbox}
\usepackage{verbatim}

\theoremstyle{plain}
\newtheorem{theorem}{Theorem}[section]

\newtheorem{lemma}[theorem]{Lemma}

\theoremstyle{definition}

\theoremstyle{remark}

\renewcommand{\algorithmicrequire}{\textbf{Input:}} 
\renewcommand{\algorithmicensure}{\textbf{Output:}} 

\newcommand{\ours}[0]{\textsc{Vrm}}
\newcommand{\rewardm}[0]{\textsc{Rm}}

\newcommand{\dpo}[0]{\textsc{Dpo}}
\newcommand{\ipo}[0]{\textsc{Ipo}}
\newcommand{\kto}[0]{\textsc{Kto}}
\newcommand{\simpo}[0]{\textsc{Simpo}}
\newcommand{\wpo}[0]{\textsc{Wpo}}
\newcommand{\sdpo}[0]{\textsc{Selective Dpo}}

\newcommand{\rlhf}[0]{\textsc{Rlhf}}
\newcommand{\ppo}[0]{\textsc{Ppo}}

\usepackage[capitalize,noabbrev]{cleveref}

\usepackage[textsize=tiny]{todonotes}

\icmltitlerunning{VRM: Teaching Reward Models to Understand Authentic Human Preferences}

\begin{document}

\twocolumn[
  \icmltitle{VRM: Teaching Reward Models to Understand Authentic Human Preferences}



  \icmlsetsymbol{equal}{*}

  \begin{icmlauthorlist}
    \icmlauthor{Biao Liu}{seu}
    \icmlauthor{Ning Xu}{seu}
    \icmlauthor{Junming Yang}{seu}
    \icmlauthor{Xuhao}{seu}
    \icmlauthor{Xin Geng}{seu}
  \end{icmlauthorlist}

  \icmlaffiliation{seu}{Southeast University, Nanjing, China}

  \icmlcorrespondingauthor{Ning Xu}{xning@seu.edu.cn}

  \icmlkeywords{Machine Learning, ICML}

  \vskip 0.3in
]



\printAffiliationsAndNotice{}  

\begin{abstract}
Large Language Models (LLMs) have achieved remarkable success across diverse natural language tasks, 
yet the reward models employed for aligning LLMs often encounter challenges of reward hacking,
where the approaches  predominantly rely on directly mapping prompt-response pairs to scalar scores, which may inadvertently capture spurious correlations rather than authentic human preferences.
In contrast, human evaluation employs a sophisticated process that initially weighs the relative importance of multiple high-dimensional objectives according to the prompt context, subsequently evaluating response quality through low-dimensional semantic features such as logical coherence and contextual appropriateness.
Motivated by this consideration, we propose {\ours}, i.e., Variational Reward Modeling, a novel framework that explicitly models the evaluation process of human preference judgments by incorporating both high-dimensional objective weights and low-dimensional semantic features as latent variables, which are inferred through variational inference techniques. 
Additionally, we provide a theoretical analysis showing that {\ours} can achieve a tighter generalization error bound compared to the traditional reward model.
Extensive experiments on benchmark datasets demonstrate that {\ours} significantly outperforms existing methods in capturing authentic human preferences.
\end{abstract}

\section{Introduction}
Large Language Models (LLMs) have demonstrated remarkable capabilities across a wide range of natural language processing tasks, including text generation \citep{liang2024controllable}, conversational interaction \citep{wang2023enabling}, reasoning \citep{xu2025towards}, and code completion \citep{jiang2024survey}.
However, ensuring that these models generate responses that align with human values and preferences remains a significant challenge.

Traditional approaches to aligning LLMs with human preferences often rely on Reinforcement Learning with Human Feedback (RLHF)~\citep{christiano2017deep,stiennon2020learning,Ouyang0JAWMZASR22}, which involves training a reward model to predict human preferences and subsequently fine-tuning the LLM to maximize the predicted rewards by policy gradient optimization~\citep{schulman2017proximal}.
While RLHF has achieved notable success, it is often computationally intensive and can suffer from instability during training~\citep{dong2023raft,yuan2023rrhf}.
To address these challenges, Direct Preference Optimization ({\dpo})~\citep{RafailovSMMEF23} has emerged as a promising alternative, offering a more straightforward approach that directly optimizes the LLM based on pairwise human preference data and is mathematically equivalent to RLHF under certain conditions.
Building upon the success of {\dpo}, several extensions have been proposed to enhance optimization efficiency, training stability, and alignment performance~\citep{pmlr-v235-ethayarajh24a,hong2024orpo,meng2024simpo,kimspread,garg2025ipo,yang2025alignment,deng2025less}.

However, existing reward model training methods primarily directly map prompt-responses pairs to scores, which may inadvertently capture spurious correlations rather than authentic human preferences, a phenomenon known as reward hacking~\citep{gao2023scaling,miao2024inform}.
For instance, the language models might exploit rewards by repeating key phrases or padding with irrelevant details. In contrast, human preference evaluation is not a straightforward scoring process. Instead, it begins with considering the relative importance of multiple high-dimensional objectives based on the prompt, such as prioritizing safety for prompts that could elicit harmful responses, or helpfulness for general queries. Subsequently, humans evaluate how well the response satisfies the high-dimensional objectives through analyzing the low-dimensional semantic features, such as logical coherence and contextual appropriateness, ultimately arriving at a holistic judgment. 
Therefore, we propose integrating this human expert-like evaluation process into reward models to better capture authentic human preferences.

Motivated by this observation, we introduce {\ours}, i.e., Variational Reward Modeling, a novel framework for training reward models that explicitly models the generative process of human preference judgments. 
Specifically, given a prompt, we assume that the high-dimensional objective weights follow a Dirichlet distribution, representing the relative importance of different objectives. Concurrently, the low-dimensional semantic features follow a multivariate Gaussian distribution, capturing the content of the prompt and response. Ultimately, the reward score is determined by both the high-dimensional objective weights and the low-dimensional semantic features.
By employing variational inference techniques \citep{xu2022variational,KingmaW13}, we can infer these latent variables from observed prompt-response pairs and optimize the reward model to better align with human preferences. Our contributions are summarized as follows:
\begin{enumerate}[leftmargin=*,itemsep=0pt,topsep=0pt]
    \item Practically, we propose a novel framework for reward model training that explicitly models the generative process of human preference judgments, incorporating both high-dimensional objective weights and low-dimensional semantic features.
    \item Theoretically, we provide a theoretical analysis of {\ours} framework proving that it can achieve a tighter generalization error bound compared to the traditional reward model training method.
\end{enumerate}
We conduct extensive experiments on benchmark datasets, demonstrating that {\ours} outperforms existing reward modeling techniques in terms of alignment with human preferences and overall performance.

\section{Related Work}

\textbf{Online LLM Alignment.} 
Building responsible and effective AI systems requires aligning Large Language Models (LLMs) with human values and intentions~\citep{achiam2023gpt,chen2025reasoning,yang2025qwen3}. A dominant paradigm for achieving this alignment is Reinforcement Learning with Human Feedback (RLHF)~\citep{christiano2017deep,stiennon2020learning,Ouyang0JAWMZASR22}. RLHF typically follows a two-stage pipeline: first training a reward model on pre-collected preference dataset to encode human preference signals, and then fine-tuning the LLM with a KL-regularized objective to maximize expected reward~\citep{Ouyang0JAWMZASR22}.
While early RLHF systems often rely on PPO-style actor-critic optimization~\citep{schulman2015trust,schulman2017proximal}, recent work has explored more scalable online optimizers that reduce dependence on a learned critic. In particular, \text{GRPO}~\citep{shao2024deepseekmath} computes group-relative advantages from multiple sampled responses to replace the value-function baseline, improving memory efficiency and practical training stability. More broadly, a growing body of work studies critic-free REINFORCE updates with stronger variance reduction~\citep{ahmadian2024back,hu2025reinforce++}, explores sequence-level optimization and clipping~\citep{zheng2025group}, and develops dynamics sampling methods for large-scale online RL~\citep{yu2025dapo}, aiming to improve efficiency and scalability~\citep{feng2025group,liu2025understanding,chen2025minimax}.

\textbf{Offline LLM Alignment.} 
While RLHF has demonstrated considerable success, it faces significant challenges including hyperparameter complexity and online sampling overhead~\citep{dong2023raft,yuan2023rrhf}. {\dpo}~\citep{RafailovSMMEF23} emerged as a compelling alternative that simplifies alignment by directly optimizing a closed-form objective on offline preference comparisons, yielding improved computational efficiency. Under specific assumptions, {\dpo} can be shown to recover an RLHF-style KL-regularized solution, while avoiding explicit on-policy RL optimization. This formulation has inspired a broad range of follow-up methods that refine the preference objective and training recipe to improve optimization efficiency, robustness, and empirical alignment quality~\citep{pmlr-v235-ethayarajh24a,hong2024orpo,kimspread,deng2025less,wu2024beta}, including alternative loss and regularization designs~\citep{garg2025ipo,hu2025explicit}, reduced reliance on reference models~\citep{hong2024orpo,meng2024simpo}, and improved mitigation of sample distribution shift~\cite{zhou2023lima,deng2025less,yang2025alignment}.


\section{Preliminaries}
We first introduce the notation for language model alignment and reward modeling. Consider a language model parameterized by $ \theta $  $ \pi_\theta:\mathcal{X}\rightarrow\mathcal{Y} $  that generates a response $\bm{y}$ given a prompt $\bm{x} $. The goal of alignment is to ensure that the generated response $\bm{y}$ aligns with human preferences, where the prompt $\bm{x} \in \mathcal{X}$ and response $\bm{y} \in \mathcal{Y}$ are sequences of tokens from a vocabulary $\mathcal{V}$, i.e., $\bm{x} = (x^1, x^2, \ldots, x^{m})$, $\bm{y} = (y^1, y^2, \ldots, y^{n})$, with $x^i, y^i \in \mathcal{V}$.

To achieve this, we collect human preference data in the form of pairs $(\bm{y}^+, \bm{y}^-)$ where $\bm{y}^+$ is preferred over $\bm{y}^-$ for a given prompt $\bm{x}$. 
The dataset is denoted as $\mathcal{D} = \{(\bm{x}_i, \bm{y}_i^+, \bm{y}_i^-)\}_{i=1}^N$.
A reward model $r: \mathcal{X} \times \mathcal{Y} \rightarrow \mathbb{R}$ is trained to predict the human preference scores. The reward model is typically trained using a loss function that encourages it to assign higher scores to preferred responses, such as the Bradley-Terry loss:
\begin{equation}\label{eq:rm}
\begin{aligned}
\mathcal{L}_{\text{rm}}
&= -\frac{1}{N}\sum_{i=1}^{N} \log P(\bm y_i^+ \succ \bm y_i^- \mid \bm x_i) \\
&= -\frac{1}{N}\sum_{i=1}^{N} \log \sigma\!\big(
    r(\bm x_i, \bm y_i^+)
    - r(\bm x_i, \bm y_i^-)
\big),
\end{aligned}
\end{equation}
where $\sigma(\cdot)$ is the sigmoid function.
Subsequently, the language model is fine-tuned to maximize the expected reward while maintaining proximity to a reference model $\pi_{\text{ref}}$ using a KL-regularized objective:
\begin{equation}\label{eq:rlhf}
\begin{aligned}
\mathcal{L}_{\rlhf}(\theta)
&= \mathbb{E}_{\substack{
\bm{x} \sim \mathcal{D},\\
\bm{y} \sim \pi_\theta(\cdot|\bm{x})
}}
\Big[
- r(\bm{x}, \bm{y}) \\
&\qquad\qquad
+ \beta \, \mathrm{KL}\!\left(
\pi_\theta(\cdot|\bm{x})
\,\|\, \pi_{\text{ref}}(\cdot|\bm{x})
\right)
\Big],
\end{aligned}
\end{equation}
where $\beta$ is a hyperparameter that balances reward maximization and KL regularization the reference model $\pi_{\text{ref}}$ is often chosen as a supervised fine-tuned model trained on human demonstrations.

\section{Methodology}
\begin{figure*}[htbp]
    \centering
    \includegraphics[width=0.87\textwidth]{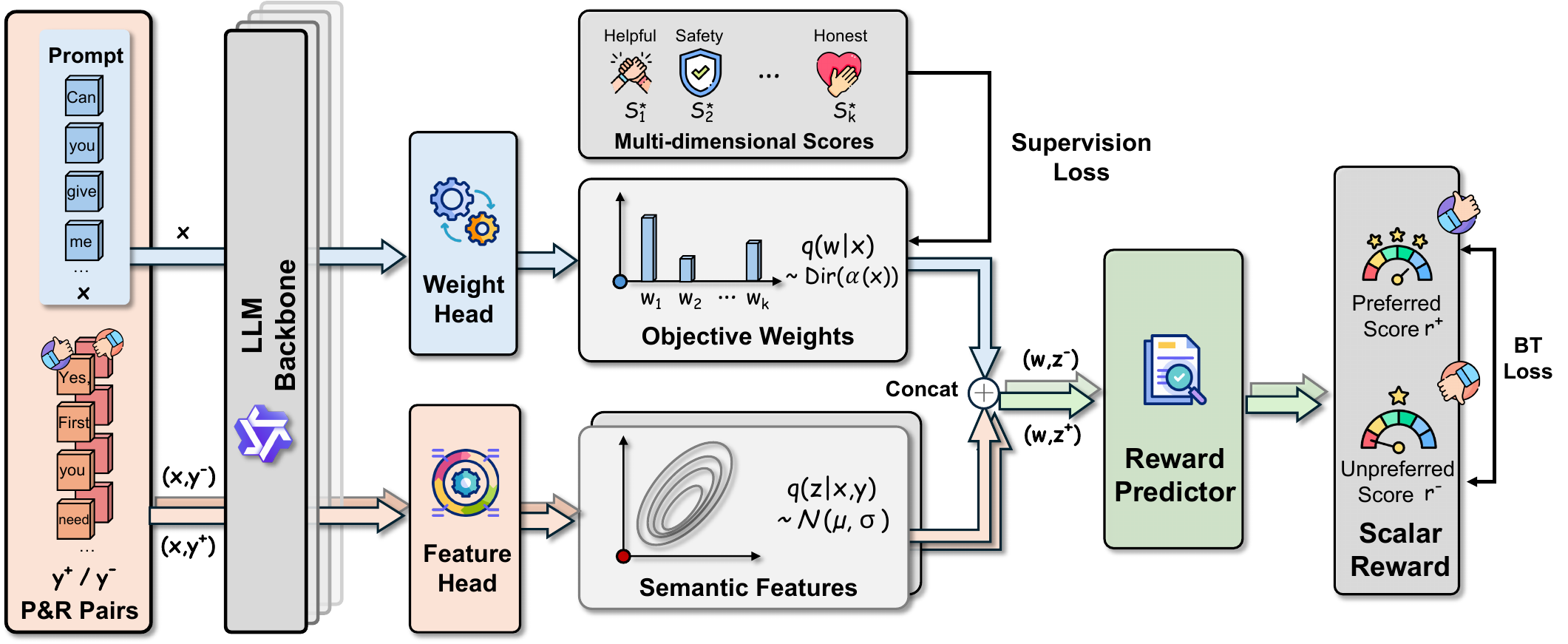}
    \caption{Overview of the {\ours} framework. The model processes prompt-response pairs through a shared backbone, generating multi-dimensional scores via the weight head and semantic features through the feature head. These components are combined with objective weights to produce the final reward predictions for preference learning.}
    \label{fig:framework}
\end{figure*}

Existing reward model training approaches typically use statistical fitting to directly map prompt-response pairs to scores, which may capture spurious correlations rather than genuine human preferences. In this section, we propose a novel framework that explicitly models the generative process of human preference judgments. The overall framework is illustrated in Figure~\ref{fig:framework}.

\subsection{Variational Reward Model}
\begin{figure}[t]
    \centering
    \includegraphics[width=0.5\linewidth]{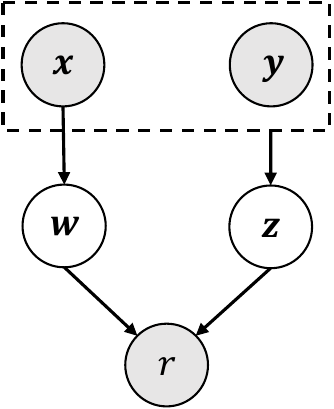}
    \caption{Causal Graph of {\ours}. The reward score $ r $ is influenced by both high-dimensional objective weights $ \bm{w} $ and low-dimensional semantic features $ \bm{z} $.}
    \label{fig:causal_rm}
\end{figure}

When presented with a prompt $\bm{x}$, we assume there exists a latent vector of objective weights $\bm{w} = [w^1, w^2, \ldots, w^K]$ that represents the relative importance of $K$ distinct objectives considered by humans when evaluating responses. For instance, for a prompt that induces the model to generate a potentially harmful response, the weight corresponding to safety would be higher. In addition to these high-dimensional objectives, the evaluation of a response also depends on low-dimensional semantic features $\bm{z} = [z^1, z^2, \ldots, z^J]$ that capture aspects such as coherence, fluency, and relevance to the prompt, where $J$ is the dimensionality of the semantic feature space. Ultimately, the reward score $ r $  is determined by both the high-dimensional objective weights and the low-dimensional semantic features. 
This generative process of human preference is illustrated in Figure~\ref{fig:causal_rm}.

Based on the causal graph depicted in Figure~\ref{fig:causal_rm}, the posterior density of the latent variables $\bm{w}$ and $\bm{z}$ can be inferred from the observed variables $(\bm{x}, \bm{y})$. Specifically, $p(\bm{w}, \bm{z} \mid \bm{x}, \bm{y})$ can be factorized as:
\begin{equation}
p(\bm{w}, \bm{z} \mid \bm{x}, \bm{y})
= p(\bm{w} \mid \bm{x}) \,p(\bm{z} \mid \bm{x}, \bm{y}),
\end{equation}
where we assume that the objective weights $\bm{w}$ depend solely on the prompt $\bm{x}$, while the semantic features $\bm{z}$ depend on both the prompt $\bm{x}$ and the response $\bm{y}$. Then $ q(\bm{w} \mid \bm{x}) $ and $ q(\bm{z} \mid \bm{x}, \bm{y}) $ can be parameterized using neural networks with parameters $\phi_1$ and $\phi_2$ to approximate the true posterior distributions of $\bm{w}$ and $\bm{z}$, respectively. 
To simplify notation, we denote all encoder parameters as $\phi = [\phi_1, \phi_2]$. 
Assume that $\bm{w}$ follows a Dirichlet distribution with parameter $\bm{\alpha} = [\alpha^1, \alpha^2, \ldots, \alpha^K]$:
\begin{equation}\label{eq:dir}
q_{\phi_1}(\bm{w} \mid \bm{x}) = \text{Dir}({\bm{\alpha}}),
\end{equation}
where ${\bm{\alpha}}$ is the output of a neural network taking $\bm{x}$ as input. Similarly, we assume that $\bm{z}$ follows a multivariate Gaussian distribution:
\begin{equation}\label{eq:gaussian}
q_{\phi_2}(\bm{z} \mid \bm{x}, \bm{y}) = \mathcal{N}(\bm{\mu}, \operatorname{diag}(\bm{\sigma}^2)),
\end{equation}
where $\bm{\mu}=[\mu^1,\mu^2,\ldots,\mu^J]$ and $\bm{\sigma}=[\sigma^1,\sigma^2,\ldots,\sigma^J]$ are outputs of another neural network taking both $\bm{x}$ and $\bm{y}$ as inputs.

Using variational Bayesian inference, we can optimize the model parameters by maximizing the Evidence Lower Bound (ELBO) to ensure that the approximate posterior distribution $q_{\phi}(\bm{w}, \bm{z} \mid \bm{x}, \bm{y})$ closely approximates the true posterior distribution $p(\bm{w}, \bm{z} \mid \bm{x}, \bm{y})$. The observed variables $\bm{x}, \bm{y}, r$ can be expressed as:
\begin{equation}
\begin{aligned}
&\log p(\bm{x}, \bm{y}, r) = \\
&\mathcal{L}_\text{ELBO}
+ \mathrm{KL}\big[
q_{\phi}(\bm{w}, \bm{z} \mid \bm{x}, \bm{y})
\,\|\, p(\bm{w}, \bm{z} \mid \bm{x}, \bm{y}, r)
\big],
\end{aligned}
\end{equation}
where the ELBO is:
\begin{equation}
\begin{aligned}
\mathcal{L}_\text{ELBO}
&= \mathbb{E}_{q_{\phi}}[\log p(r \mid \bm{w}, \bm{z})] \\
&\quad - \mathrm{KL}\big[
q_{\phi_1}(\bm{w} \mid \bm{x})
\,\|\, p(\bm{w})
\big] \\
&\quad - \mathrm{KL}\big[
q_{\phi_2}(\bm{z} \mid \bm{x}, \bm{y})
\,\|\, p(\bm{z})
\big].
\end{aligned}
\end{equation}
The detailed derivation of the ELBO is provided in Appendix~\ref{app:elbo}. Due to the non-negativity of the KL divergence and $ \log p(\bm x, \bm y, r) $ is a constant, maximizing the ELBO is equivalent to minimizing the KL divergence between the approximate and true posterior distributions.
The prior distribution of the objective weights $\bm{w}$ is assumed to be a Dirichlet distribution with parameter $\bm{\alpha}_0 = [\alpha_0^1, \alpha_0^2, \ldots, \alpha_0^K]$, allowing us to compute the second term of the ELBO as:
\begin{equation}
\begin{aligned}
&\mathrm{KL}\big[
q_{\phi_1}(\bm{w} \mid \bm{x})
\,\|\, p(\bm{w})
\big] = 
\log \Gamma(\sum_k \alpha^k) \\
&- \sum_{k} \log \Gamma({\alpha}^k)
- \log \Gamma(\sum_k \alpha_0^k) + \sum_{k} \log \Gamma(\alpha_0^k) \\
&+ \sum_{k} ({\alpha}^k - \alpha_{0}^k) \big[
\Psi({\alpha}^k) -  \Psi(\sum_{k} {\alpha}^k)
\big],
\end{aligned}
\end{equation}
where $\Gamma(\cdot)$ is the gamma function, and $\Psi(\cdot)$ is the digamma function.
Similarly, assuming that the prior distribution of the semantic features $\bm{z}$ is a multivariate Gaussian distribution with mean $\bm{0}$ and covariance matrix $\mathbf{I}$, we can compute the third term of the ELBO as:
\begin{equation}
\begin{aligned}
&\mathrm{KL}\big[
q_{\phi_2}(\bm{z} \mid \bm{x}, \bm{y})
\,\|\, p(\bm{z})
\big] = \\
&\frac{1}{2} \sum_{j} \bigg( (\mu^j)^2 + (\sigma^j)^2 - 2 \log \sigma^j - 1 \bigg).
\end{aligned}
\end{equation}

The first term of the ELBO, $\mathbb{E}_{q_{\phi}}[\log p(r \mid \bm{w}, \bm{z})]$, can be computed using Monte Carlo sampling by drawing samples of $\bm{w}$ and $\bm{z}$ from their respective approximate posterior distributions. To preserve gradient information through the sampling process, we employ the reparameterization trick~\citep{figurnov2018implicit}. Assuming $M$ samples are drawn, the first term of the ELBO can be approximated as:
\begin{equation}
\begin{aligned}
&\mathbb{E}_{q_{\phi}}[\log p(r \mid \bm{w}, \bm{z})] \approx \\
&\frac{1}{M} \sum_{m=1}^{M} \log p(r \mid \bm{w}^{(m)}, \bm{z}^{(m)}),
\end{aligned}
\end{equation}
where $\bm{w}^{(m)}$ and $\bm{z}^{(m)}$ are samples drawn from $q_{\phi_1}(\bm{w} \mid \bm{x})$ and $q_{\phi_2}(\bm{z} \mid \bm{x}, \bm{y})$, respectively. In practice, we use a single sample ($M=1$) to estimate this expectation as suggested by \citet{KingmaW13}.

For preference datasets $\mathcal{D} = \{(\bm{x}_i, \bm{y}_i^+, \bm{y}_i^-)\}_{i=1}^N$, we can extend the ELBO formulation to accommodate pairwise comparisons. The modified ELBO becomes:
\begin{equation}\label{eq:pref_elbo}
\begin{aligned}
\mathcal{L}_\text{ELBO} &= \frac{1}{N} \sum_{i=1}^{N} \mathbb{E}_{q_{\phi}} \big[\log P(\bm{y}_i^+ \succ \bm{y}_i^- \mid \bm{x}_i)\big] \\
&\quad - \mathrm{KL}\big[
q_{\phi_1}(\bm{w} \mid \bm{x}_i)
\,\|\, p(\bm{w})
\big] \\
&\quad - \mathrm{KL}\big[
q_{\phi_2}(\bm{z} \mid \bm{x}_i, \bm{y}_i^+)
\,\|\, p(\bm{z})
\big] \\
&\quad - \mathrm{KL}\big[
q_{\phi_2}(\bm{z} \mid \bm{x}_i, \bm{y}_i^-)
\,\|\, p(\bm{z})
\big].
\end{aligned}
\end{equation}
Here, $P(\bm{y}_i^+ \succ \bm{y}_i^- \mid \bm{x}_i)$ represents the probability that response $\bm{y}_i^+$ is preferred over $\bm{y}_i^-$ given the prompt $\bm{x}_i$. This probability can be modeled using the Bradley-Terry framework:
\begin{equation}
\begin{aligned}
&P(\bm{y}_i^+ \succ \bm{y}_i^- \mid \bm{x}_i) = \\
&\sigma\big(
r_{\psi}(\bm w, \bm z^+)
- r_{\psi}(\bm w, \bm z^-)
\big), 
\end{aligned}
\end{equation}
where $r_{\psi}(\bm{w}, \bm{z})$ is the reward score computed based on the latent variables $\bm{w}$, $ \bm z^+ $ and $ \bm z^- $. The parameters of the reward score decoder are denoted as $\psi$.

\subsection{Supervision of Objective Weights}
To constrain the latent variable $\bm{w}$, we introduce a supervision term to ensure that $\bm{w}$ captures the high-dimensional objective weight information of the prompt. Some datasets provide multi-dimensional scores for each prompt and its corresponding responses, such as Helpful, Honest, Harmless~\citep{cui2023ultrafeedback,ji-etal-2025-pku}. We can leverage these multi-dimensional scores to supervise the learning of the latent variable $\bm{w}$. Specifically, for each prompt $\bm{x}_i$ and responses $\bm{y}_i^+$ and $\bm{y}_i^-$, we have corresponding multi-dimensional score vectors $\bm{s}_i^+ = [s_i^{1,+}, s_i^{2,+}, \ldots, s_i^{K,+}]$ and $\bm{s}_i^- = [s_i^{1,-}, s_i^{2,-}, \ldots, s_i^{K,-}]$. 
The scores of the preferred responses implicitly contain the expert's preference weights during evaluation.
We normalize the scores for each dimension using mean-variance normalization and then apply the softmax function to convert them into a probability distribution, which serves as the supervision signal for $\bm{w}$. Specifically, we define the supervision loss as:
\begin{equation}
\begin{aligned}
\mathcal{L}_\text{sup} &= \frac{1}{N} \sum_{i=1}^{N}
\mathrm{KL}\big[
q_{\phi_1}(\bm{w} \mid \bm{x}_i)
\,\|\, \tilde{\bm{s}}_i^+
\big],
\end{aligned}
\end{equation}
where $\tilde{\bm{s}}_i^+$ is the normalized score vector for response $\bm{y}_i^+$.

The overall training objective for {\ours} combines the ELBO and the supervision loss:
\begin{equation}\label{eq:final_loss}
\mathcal{L} = -\mathcal{L}_\text{ELBO} + \lambda \mathcal{L}_\text{sup},
\end{equation}
where $\lambda$ is a hyperparameter that balances the two components. 

\section{Theoretical Analysis}
\subsection{A PAC-Bayes Generalization Bound}
We derive a generalization error bound for our variational reward model based on the PAC-Bayesian model averaging theorem~\citep{mcallester1999pacbayesian}. Consider a preference dataset
$\mathcal{S}=\{(\bm{x}_i,\bm{y}_i^+,\bm{y}_i^-)\}_{i=1}^{N}$ drawn i.i.d. from an unknown distribution $\mathcal{D}$.
Given latent variables, our reward model predicts a pairwise preference by comparing
$r_\psi(\bm{w},\bm{z}^+)$ and $r_\psi(\bm{w},\bm{z}^-)$.
We analyze the $0$-$1$ preference error:
\begin{equation}
\ell(\bm{x},\bm{y}^+,\bm{y}^-)
=\mathbb{I}\!\left[r_\psi(\bm{w},\bm{z}^+) \le r_\psi(\bm{w},\bm{z}^-)\right],
\label{eq:pref_01_loss}
\end{equation}
where $\mathbb{I}[\cdot]$ is the indicator function. Our inference networks define a data-dependent posterior over latent variables:
\begin{equation}
\begin{aligned}
&q_{\phi}(\bm{w},\bm{z}^+,\bm{z}^- \mid \bm{x},\bm{y}^+,\bm{y}^-)
= q_{\phi_1}(\bm{w}\mid \bm{x})\,\\
&q_{\phi_2}(\bm{z}^+\mid \bm{x},\bm{y}^+)\,
q_{\phi_2}(\bm{z}^-\mid \bm{x},\bm{y}^-),
\end{aligned}
\end{equation}
and we use a factorized prior $p(\bm{w},\bm{z}^+,\bm{z}^-)=p(\bm{w})p(\bm{z}^+)p(\bm{z}^-)$.
Define the population and empirical risks:
\begin{equation}
\mathcal{R} = \mathbb{E}_{(\bm{x},\bm{y}^+,\bm{y}^-)\sim \mathcal{D}}\;
\mathbb{E}_{q_{\phi}(\bm{w},\bm{z}^+,\bm{z}^- \mid \bm{x},\bm{y}^+,\bm{y}^-)}\!\left[\ell\right],
\end{equation}
\begin{equation}
\widehat{\mathcal{R}} = \frac{1}{N}\sum_{i=1}^{N}
\mathbb{E}_{q_{\phi}(\bm{w},\bm{z}_i^+,\bm{z}_i^- \mid \bm{x}_i,\bm{y}_i^+,\bm{y}_i^-)}\!\left[\ell\right].
\end{equation}

\begin{theorem}[PAC-Bayes bound with decomposed KL]\label{thm:pacbayes_vrm}
With probability at least $1-\delta$ over the draw of the training sample $\mathcal{S}$, the following holds:
\begin{equation}
\begin{aligned}
&\mathcal{R} \le \widehat{\mathcal{R}}
+ \Bigg(\frac{1}{2N-1}\!\Big(
D(Q \| P)
\\
& + \ln\frac{1}{\delta} + \frac{5}{2}\ln N + 8
\Big)\Bigg)^{1/2},
\end{aligned}
\end{equation}
where $D(Q \| P) $ is the total KL divergence over the training set:
\begin{equation}
\begin{aligned}
D(Q \| P)
&= \sum_{i=1}^{N} \Big(
\mathrm{KL}\!\left(q_{\phi_1}(\bm{w}\mid \bm{x}_i)\,\|\,p(\bm{w})\right) \\
& + \mathrm{KL}\!\left(q_{\phi_2}(\bm{z}\mid \bm{x}_i,\bm{y}_i^+)\,\|\,p(\bm{z})\right) \\
& + \mathrm{KL}\!\left(q_{\phi_2}(\bm{z}\mid \bm{x}_i,\bm{y}_i^-)\,\|\,p(\bm{z})\right)
\Big).
\end{aligned}
\end{equation}
\end{theorem}
The bound indicates that with a sufficiently large training sample, minimizing the empirical error leads to a tight upper bound on the population preference error, with a small complexity penalty. 
In contrast, directly inputting prompt-response pairs into a single reward model without considering latent variables and intermediate decision processes results in a fixed KL divergence that cannot be optimized, leading to a looser generalization error bound.

\section{Experiments}

\begin{table*}[!t]
\centering
\caption{Performance comparison across AlpacaEval 2, Arena-Hard, and MT-Bench benchmarks.}
\resizebox{\linewidth}{!}{%
\begin{tabular}{lcccccccccc}
\toprule
\multirow{3}{*}{\textbf{Methods}} 
& \multicolumn{5}{c}{\textbf{Qwen2.5-7B}} 
& \multicolumn{5}{c}{\textbf{Qwen 3-8B}} \\
\cmidrule(lr){2-6} \cmidrule(lr){7-11}
& \multicolumn{2}{c}{\textbf{AlpacaEval 2}} 
& \multicolumn{2}{c}{\textbf{Arena-Hard}} 
& \textbf{MT-Bench}
& \multicolumn{2}{c}{\textbf{AlpacaEval 2}} 
& \multicolumn{2}{c}{\textbf{Arena-Hard}} 
& \textbf{MT-Bench} \\
\cmidrule(lr){2-3} \cmidrule(lr){4-5} \cmidrule(lr){6-6} \cmidrule(lr){7-8} \cmidrule(lr){9-10} \cmidrule(lr){11-11}
& WR(\%) & LC(\%) & WR(\%) & SC(\%) & Score
& WR(\%) & LC(\%) & WR(\%) & SC(\%) & Score \\
\midrule
{\dpo}      & 37.24 & 36.84 & 49.0 & 47.2 & 7.83 & 18.53 & 31.50 & 70.8 & 71.9 & 7.86 \\
{\ipo}      & 37.95 & 36.43 & 54.6 & 48.3 & 7.64 & 27.79 & 25.87 & 65.5 & 54.9 & 6.55 \\
{\kto}      & 38.12 & 36.51 & 43.9 & 44.1 & 7.63 & 22.94 & 39.68 & 69.5 & 70.1 & 7.97 \\
{\simpo}    & 40.03 & 40.78 & 54.6 & 48.8 & 7.58 & 29.38 & 27.94 & 61.9 & 56.6 & 6.24 \\
{\wpo}      & 44.11 & 40.06 & 62.0 & 53.0 & 7.81 & 28.16 & 25.35 & 56.2 & 53.7 & 6.38 \\
{\sdpo}     & 38.02 & 39.21 & 51.7 & 48.2 & 7.74 & 29.27  & 27.53 & 60.4 & 51.8 & 6.13 \\
{\ppo}      & 39.52 & 39.79 & 55.3 & 48.9 & 7.81 & 44.75 & \textbf{55.33} & 60.4 & 71.9 & 8.44 \\
\midrule
{\ours}-{\ppo}
            & \textbf{45.30} & \textbf{50.38} & \textbf{65.5} & \textbf{57.2} & \textbf{7.98}
            & \textbf{48.14} & 52.03 & \textbf{84.0} & \textbf{82.0} & \textbf{8.58} \\
\bottomrule
\end{tabular}%
}
\label{tab:benchmark_results}
\end{table*}

\begin{table*}[thbp]
\centering
\tiny
\caption{Reward model performance on Reward-Bench and UltraFeedback-Cleaned.}
\resizebox{0.7\linewidth}{!}{%
\begin{tabular}{lccccc}
\toprule
\multirow{2}{*}{\textbf{Methods}}
& \multicolumn{4}{c}{\textbf{Reward-Bench}}
& \multicolumn{1}{c}{\textbf{UF-C}} \\
\cmidrule(lr){2-5} \cmidrule(lr){6-6}
& Chat & Chat Hard & Safety & Reasoning & Total \\
\midrule
{\dpo}   & 96.09 & 54.17 & 67.84 & 57.18 & 75.71 \\
{\ipo}   & 55.31 & 72.15 & 68.24 & 84.44 & 60.69 \\
{\kto}   & 95.25 & 43.86 & 64.86 & 54.74 & 70.67 \\
{\simpo} & 54.47 & 63.60 & 51.89 & 72.88 & 59.16 \\
{\wpo}   & 53.07 & 55.04 & 58.11 & 62.20 & 54.74 \\
{\sdpo}  & 80.97 & 84.97 & 60.53 & 82.03 & 82.28 \\

{\rewardm} &  94.04   & 93.02 & 76.66   & 86.62   & 88.98  \\
\midrule
{\ours} &  \textbf{97.11}  &   \textbf{94.23}   &   \textbf{85.41}   &   \textbf{89.72}  &   \textbf{92.36}  \\
\bottomrule
\end{tabular}%
}
\label{tab:rewardbench_ultrafeedback}
\end{table*}

\begin{figure*}[t]
\centering
\begin{subfigure}[t]{0.32\textwidth}
    \centering
    \includegraphics[width=\linewidth]{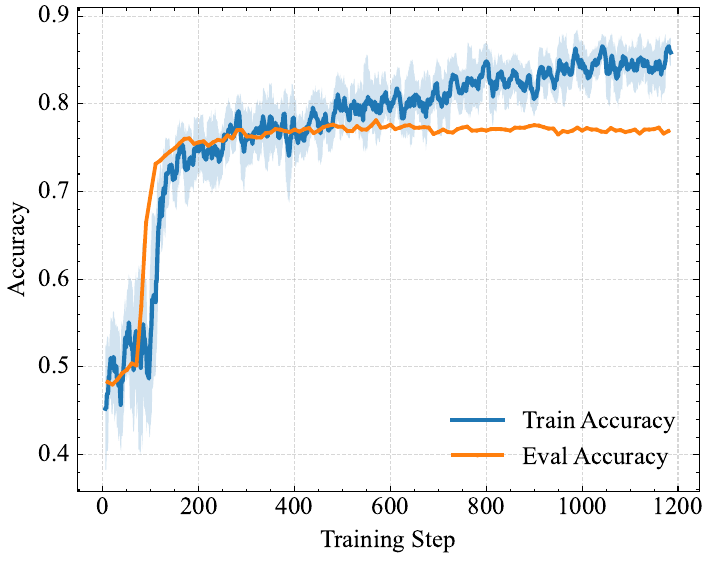}
    \caption{KL}
\end{subfigure}\hfill
\begin{subfigure}[t]{0.32\textwidth}
    \centering
    \includegraphics[width=\linewidth]{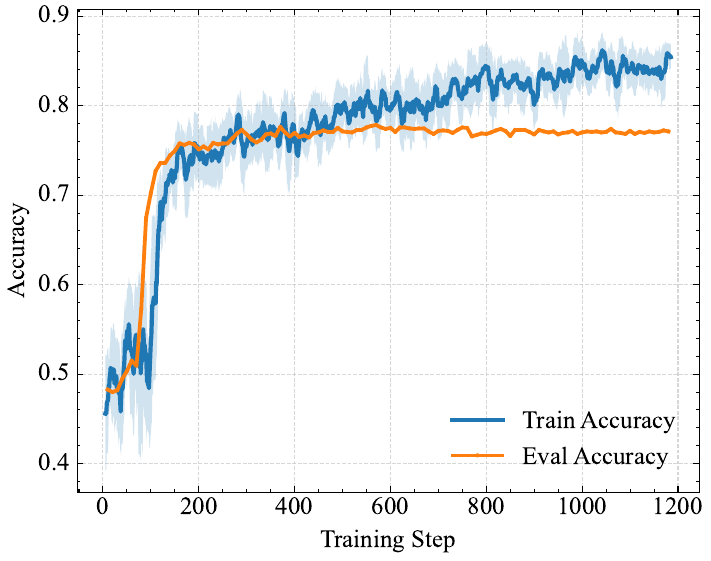}
    \caption{MAE}
\end{subfigure}\hfill
\begin{subfigure}[t]{0.32\textwidth}
    \centering
    \includegraphics[width=\linewidth]{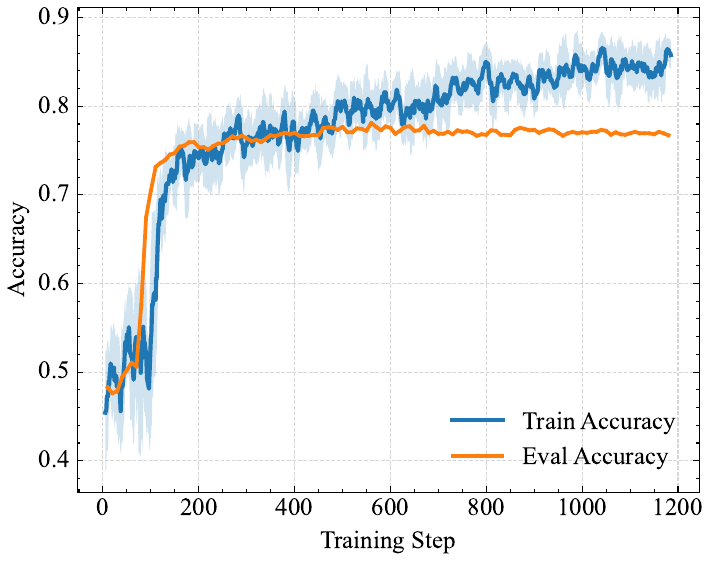}
    \caption{Rank}
\end{subfigure}
\caption{Accuracy curves under different supervision loss types.}
\label{fig:loss_type_accuracy_curves}
\end{figure*}

\begin{figure*}[t]
\centering
\begin{subfigure}[t]{0.32\textwidth}
    \centering
    \includegraphics[width=\linewidth]{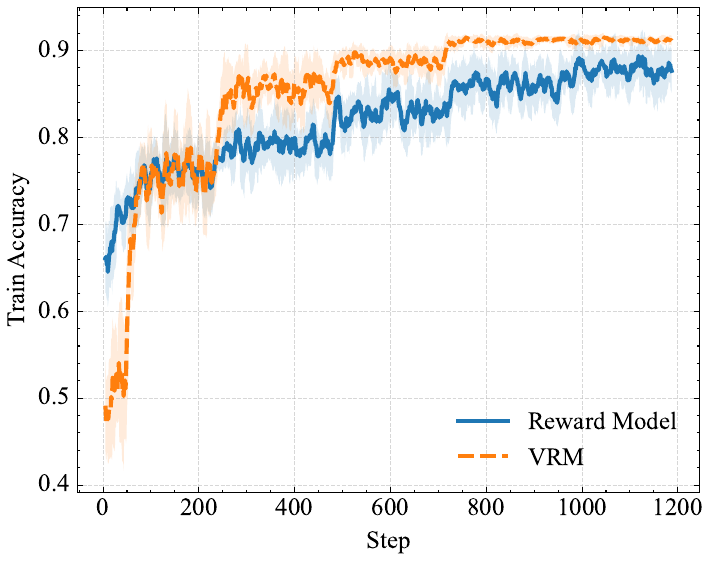}
    \caption{Train}
\end{subfigure}
\qquad
\begin{subfigure}[t]{0.32\textwidth}
    \centering
    \includegraphics[width=\linewidth]{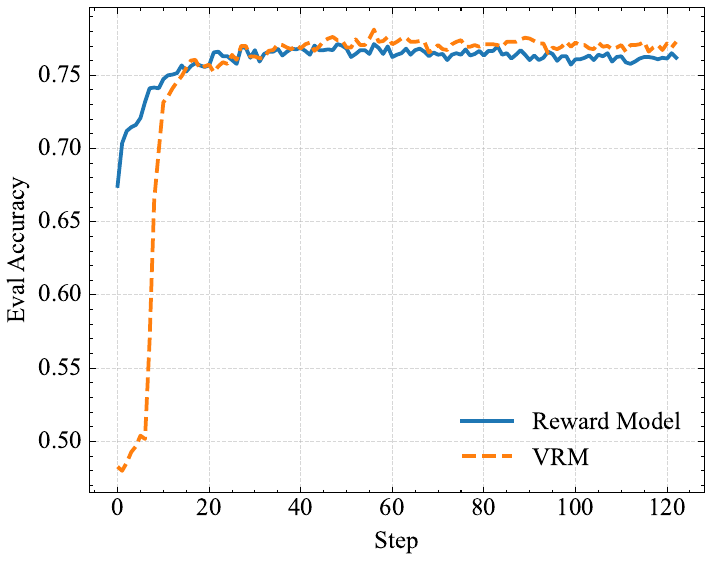}
    \caption{Eval}
\end{subfigure}
\caption{Accuracy curves comparing the traditional reward model (RM) and {\ours}.}
\label{fig:rm_vs_crm}
\end{figure*}

\begin{figure*}[t]
    \centering
    \begin{subfigure}[t]{0.32\textwidth}
        \centering
        \includegraphics[width=\linewidth]{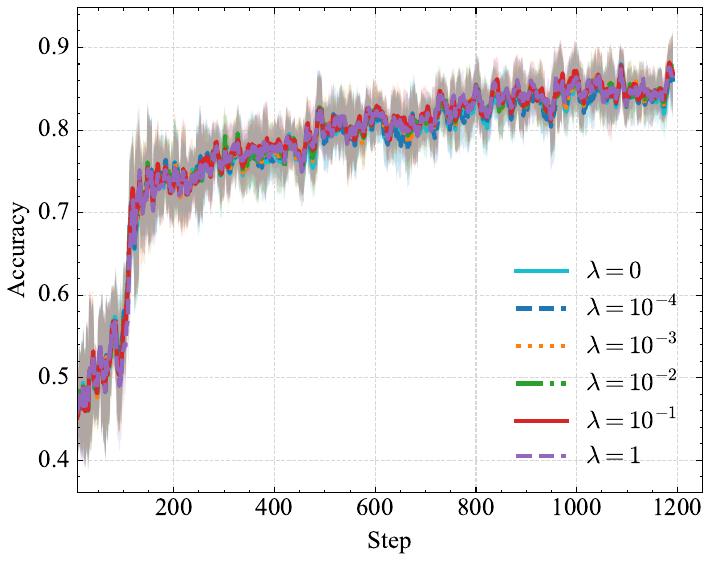}
        \caption{Train Accuracy}
    \end{subfigure}\hfill
    \begin{subfigure}[t]{0.32\textwidth}
        \centering
        \includegraphics[width=\linewidth]{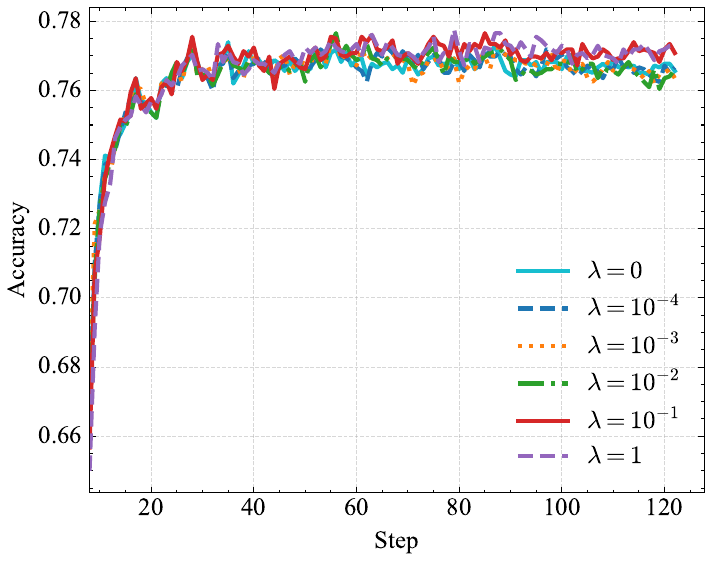}
        \caption{Eval Accuracy}
    \end{subfigure}\hfill
    \begin{subfigure}[t]{0.32\textwidth}
        \centering
        \includegraphics[width=\linewidth]{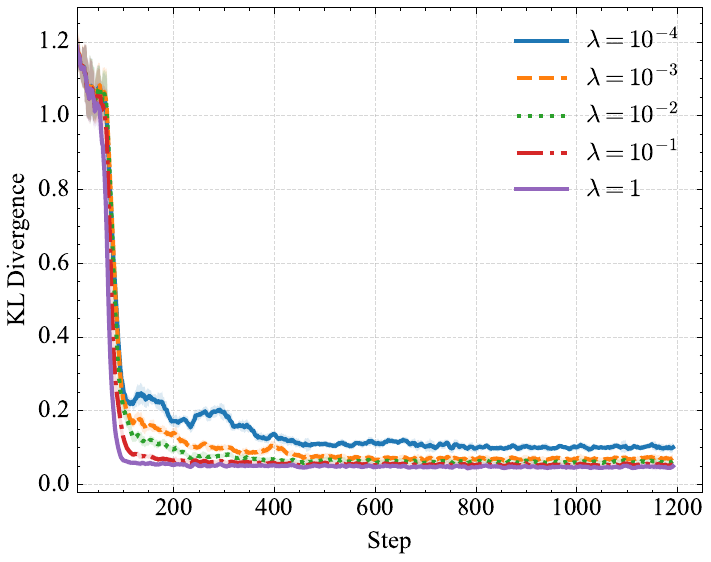}
        \caption{KL Divergence}
    \end{subfigure}
    \caption{Parameter sensitivity analysis of the supervision loss weight $\lambda$.}
    \label{fig:sup_weight_sensitivity}
    \end{figure*}

\begin{figure}[t]
    \centering
    \includegraphics[width=0.9\linewidth]{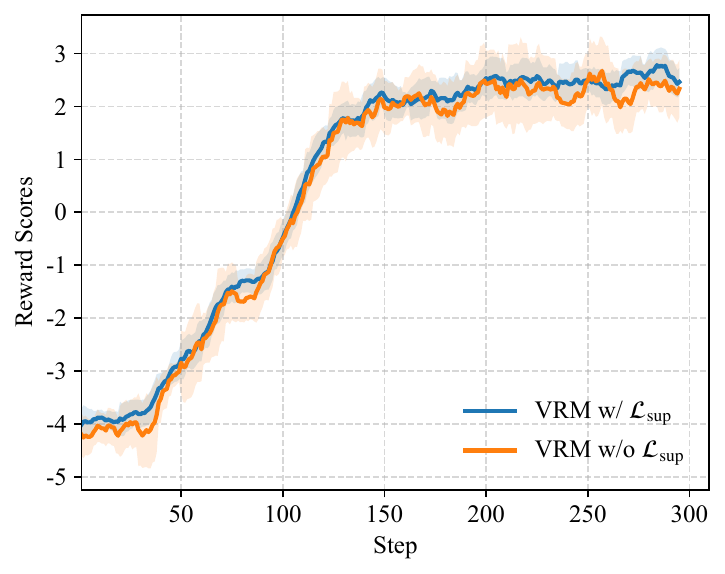}
    \caption{Reward curves comparing VRM with and without $\mathcal{L}_{\text{sup}}$.}
    \label{fig:reward_sup_vs_no_sup}
\end{figure}

\subsection{Experimental Setup}

\textbf{Datasets and Models.} We use the UltraFeedback dataset, which contains 64K annotations constructed from four LLM-generated responses per instruction and annotated by GPT-4. The dataset provides fine-grained scalar scores along multiple evaluation dimensions, including instruction-following, truthfulness, honesty, and helpfulness~\citep{cui2023ultrafeedback}. 
For the reward model implementation, we employ Qwen3-4B-Instruct-2507~\citep{yang2025qwen3} as the base model. For downstream alignment experiments, we utilize a supervised fine-tuned version of Qwen-2.5-7B\footnote{\href{https://huggingface.co/FuseAI/FuseChat-Qwen-2.5-7B-SFT}{FuseAI/FuseChat-Qwen-2.5-7B-SFT}} and Qwen3-8B~\citep{yang2025qwen3} as the base policy models to evaluate the effectiveness of our reward modeling approach in practical RLHF scenarios.

\textbf{Evaluation Metrics.} We evaluate the aligned language models using three widely adopted automatic evaluation benchmarks: AlpacaEval~\citep{alpaca_eval,dubois2024length}, Arena-Hard~\citep{li2024crowdsourced}, and MT-Bench ~\citep{bai2024mt}. For AlpacaEval 2, we report both the raw win rate (WR) and the length-controlled win rate (LC) when compared against the reference model GPT-4o-05-13 for Qwen-2.5-7B and GPT-4-Preview-1106 for Qwen3-8B. On Arena-Hard, we evaluate performance using the win rate (WR) and the style-controlled win rate (SC), with GPT-4-Preview-1106 as the baseline for all base models. For MT-Bench, we report the average multi-turn score (Score) assigned by GPT-4o, where each response is evaluated on a 10-point scale. All evaluations are conducted using GPT-4o as the judge model.
For reward model evaluation, we utilize the Reward-Bench~\citep{lambert2024rewardbenchevaluatingrewardmodels} and UltraFeedback-Cleaned\footnote{\href{https://huggingface.co/datasets/allenai/ultrafeedback_binarized_cleaned}{allenai/ultrafeedback\_binarized\_cleaned}} (UF-C) datasets. Reward-Bench consists of four categories: Chat, Chat Hard, Safety, and Reasoning, while UF-C is a cleaned subset of the UltraFeedback dataset with high-quality annotations. We report accuracy on pairwise comparisons for both datasets. 

\textbf{baselines.} we compared our method with various advanced LLM alignment methods: 
{\dpo} \citep{RafailovSMMEF23}, {\ipo} \citep{azar2024general}, {\kto} \citep{pmlr-v235-ethayarajh24a}, {\simpo} \citep{meng2024simpo}, {\wpo} \citep{ZhouAZIZSXZ24}, {\sdpo} \citep{gao2025principled}, and {\ppo} \citep{Ouyang0JAWMZASR22}. The detailed descriptions of these methods are as follows:
\begin{enumerate}[leftmargin=*, itemsep=0pt, parsep=0pt, topsep=0pt]
    \item {\dpo}: A preference-based alignment method that directly fine-tunes the model on pairwise human preferences to increase the probability of preferred responses without training an explicit reward model.
    \item {\ipo}: A variant of DPO that introduces a modified loss or regularization to reduce overfitting and improve robustness by leveraging the model’s own preference classification capabilities.
    \item {\kto}: An alignment objective inspired by prospect theory that optimizes model outputs based on binary desirability signals by directly maximizing utility rather than preference likelihood, potentially without paired preference data.
    \item {\simpo}: An efficient extension of preference optimization that uses the average log probability of sequences as an implicit reward and removes reliance on a reference model, improving compute and memory efficiency.
    \item {\wpo}: A family of extensions to preference optimization that incorporate weighting or utility adjustments to better capture differing reward magnitudes.
    \item {\sdpo}: A DPO variant that selectively focuses optimization on high-impact preference pairs or subsets to improve alignment quality and stability by prioritizing informative feedback.
    \item {\ppo}: A classic reinforcement learning method used in RLHF that updates model policies by maximizing expected reward with a clipped surrogate objective to ensure stable, constrained changes. 
\end{enumerate}
\subsection{Main Results}
The performance comparison across AlpacaEval 2, Arena-Hard, and MT-Bench benchmarks is presented in Table~\ref{tab:benchmark_results}. Our method, {\ours}-{\ppo}, consistently outperforms all baseline methods across both Qwen2.5-7B and Qwen3-8B models, expect for a slight underperformance compared to {\ppo} on Length-Controlled win rate on AlpacaEval 2 with Qwen3-8B. Notably, on Qwen2.5-7B, our method achieves a win rate of 50.38\% on LC win rate of AlpacaEval 2, surpassing the strongest baseline {\simpo} by a clear margin of over 9.6 percentage points. Similar trends are observed on Arena-Hard, where {\ours}-{\ppo} attains the highest win rate and style-controlled win rate among all methods, suggesting improved robustness on challenging prompts. 
On the larger Qwen3-8B model, {\ours}-{\ppo} continues to demonstrate strong and stable performance, achieving the best overall results on Arena-Hard and MT-Bench, with notable margins over {\ppo} and other preference-optimization baselines. While {\ppo} slightly outperforms our method on the length-controlled metric of AlpacaEval 2, {\ours}-{\ppo} still achieves the highest overall win rate.

Table~\ref{tab:rewardbench_ultrafeedback} summarizes reward model accuracy on Reward-Bench and UltraFeedback-Cleaned (UF-C). Our method (\ours) achieves the best overall performance across all Reward-Bench categories, and also obtains the highest UF-C accuracy. Compared with the strongest baseline reward model (\rewardm), {\ours} improves UF-C by 3.38 points (88.98 $\rightarrow$ 92.36) while consistently increasing category-wise accuracy in Reward-Bench, including Chat, Chat Hard, Safety, and Reasoning.
This indicates that our variational formulation, yields a reward function that generalizes better to safety- and reasoning-critical comparisons rather than overfitting to surface-level chat preferences.

\subsection{Further Analysis}

\subsection{Comparison with Traditional Reward Model}

To investigate the performance of our method compared to the traditional reward model, we compare the accuracy curves of our method and the traditional reward model. The experimental results are shown in Figure~\ref{fig:rm_vs_crm}, where we present the accuracy curves for our method and the traditional reward model.

\subsection{Loss Type of Supervision}
To investigate the impact of different supervision loss functions, we compare the use of Kullback-Leibler divergence (KL), mean squared error (MAE), and ranking loss (RANK) as the training objectives for the reward model. The experimental results are shown in Figure~\ref{fig:loss_type_accuracy_curves}, where we present the training and evaluation accuracy curves for each loss as a function of training steps. As can be observed, our method achieves very similar accuracy across all three loss types, with comparable convergence speed and final evaluation performance. This suggests that our approach is robust to the choice of supervision loss, provided the objective is reasonable.

\subsection{Parameter Sensitivity}
To investigate the sensitivity of the weight $\lambda$ for the supervision loss, we vary the value of $\lambda$ from the set $ \{1e-4, 1e-3, 1e-2, 1e-1, 1\} $ with a fixed learning rate $1e-6$. The experimental results are shown in Figure~\ref{fig:sup_weight_sensitivity}, where we present the train accuracy, eval accuracy, and KL divergence curves for different values of $\lambda$.
The train accuracy curves are almost identical for different values of $\lambda$, indicating that the training process is relatively robust to the choice of $\lambda$. The eval accuracy curves show a slight improvement with increasing $\lambda$, especially in the later plateau stage, where the curves for larger $\lambda$ are higher, indicating that stronger supervision can bring some generalization benefits. The KL divergence curves show a faster decrease and earlier stabilization with increasing $\lambda$, indicating that stronger supervision signals can accelerate the convergence of the KL constraint. Overall, increasing the supervision weight does not significantly change the training set fitting (train acc is almost the same), but can improve the validation performance and accelerate the convergence of the KL supervision term, indicating that explicit supervision can bring benefits to the training framework.

\subsection{Ablation Study}
We ablate the additional supervision term by setting $\lambda=0$. As shown in Figure~\ref{fig:sup_weight_sensitivity} (a,b), removing this term does not lead to a noticeable drop in reward-model accuracy, suggesting that our variational formulation can already learn higher-order factors to some extent without explicit supervision. The main benefit of introducing the supervision term is improved interpretability: it anchors each dimension to a human-defined high-level attribute, enabling more transparent, dimension-wise analysis of what the reward model has captured.
To further investigate the role of the supervision term, we compare the results of alignment experiments with and without this term, as shown in Figure~\ref{fig:reward_sup_vs_no_sup}. The results indicate that the supervised loss term brings some stability to the training process. However, even without this supervision term, the performance degradation is not significant, suggesting that our training framework can still automatically discover higher-order features.

\section{Conclusion}
In this paper, we proposed a novel variational reward modeling approach that leverages latent variable inference to disentangle objective weights and semantic features from prompts and responses. By incorporating a supervision mechanism based on multi-dimensional scores, our method effectively captures the underlying preferences in human feedback data.
Extensive experiments on multiple benchmarks demonstrated the superiority of our approach over existing reward modeling techniques, leading to improved alignment of large language models with human values.

\section*{Impact Statement}
The proposed variational reward modeling approach has the potential to significantly enhance the alignment of large language models with human values and preferences. By effectively capturing the underlying factors that drive human feedback, our method can lead to the development of AI systems that are more responsive to user needs and ethical considerations. However, it is essential to consider the ethical implications of deploying such models, including issues related to bias, fairness, and transparency. Future work should focus on addressing these challenges to ensure that the benefits of improved reward modeling are realized in a responsible and equitable manner.

\nocite{langley00}

\bibliography{example_paper}
\bibliographystyle{icml2026}

\newpage
\appendix
\onecolumn

\section{Detailed Derivation of ELBO}\label{app:elbo}

This section provides the complete mathematical derivation of the Evidence Lower Bound (ELBO) for the causal reward model.


By the law of total probability:
\begin{equation}
\log p(r, \bm{x}, \bm{y}) = \log p(\bm{w},\bm{z}, r, \bm{x}, \bm{y}) - \log p(\bm{w},\bm{z}\mid r, \bm{x}, \bm{y}).
\end{equation}

Multiply both sides by $q_{\theta}(\bm{w}, \bm{z}| r, \bm{x}, \bm{y})$, and for $\bm{z}$ and $\bm{w}$ integral:

\begin{equation}
\begin{aligned}
&\int_{\bm{w},\bm{z}} q_{\theta}(\bm{w}, \bm{z} | r, \bm{x}, \bm{y}) \log p(r, \bm{x}, \bm{y})\mathrm{d}\bm{w}\mathrm{d}\bm{z} \\
&\quad = \int_{\bm{w},\bm{z}} q_{\theta}(\bm{w}, \bm{z} | r, \bm{x}, \bm{y}) (\log p(\bm{w},\bm{z}, r, \bm{x}, \bm{y}) - \log p(\bm{w},\bm{z} | r, \bm{x}, \bm{y}))\mathrm{d}\bm{w}\mathrm{d}\bm{z}.
\end{aligned}
\end{equation}

On the left side, $\log p(r, \bm{x}, \bm{y})$ is independent of $\bm{z}$ and $\bm{w}$:

\begin{equation}
\begin{aligned}
\log p(r, \bm{x}, \bm{y}) &= \int_{\bm{w},\bm{z}} q_{\theta}(\bm{w}, \bm{z} | r, \bm{x}, \bm{y})(\log p(\bm{w}, \bm{z}, r, \bm{x}, \bm{y}) \nonumber - \log p(\bm{w}, \bm{z} | r, \bm{x}, \bm{y}))\mathrm{d}\bm{w}\mathrm{d}\bm{z} \nonumber \\
&= \int_{\bm{w},\bm{z}} q_{\theta}(\bm{w}, \bm{z} | r, \bm{x}, \bm{y}) \left(\log \frac{p(\bm{w}, \bm{z}, r, \bm{x}, \bm{y})}{q_{\theta}(\bm{w}, \bm{z} | r, \bm{x}, \bm{y})} - \log \frac{p(\bm{w}, \bm{z} | r, \bm{x}, \bm{y})}{q_{\theta}(\bm{w}, \bm{z} | r, \bm{x}, \bm{y})}\right)  \\
&= \int_{\bm{w},\bm{z}} q_{\theta}(\bm{w}, \bm{z} | r, \bm{x}, \bm{y})\log \frac{p(\bm{w}, \bm{z}, r, \bm{x}, \bm{y})}{q_{\theta}(\bm{w}, \bm{z} | r, \bm{x}, \bm{y})}\mathrm{d}\bm{w}\mathrm{d}\bm{z} \nonumber \\
&\quad + \mathrm{KL}[q_{\theta}(\bm{w}, \bm{z} | r, \bm{x}, \bm{y})||p(\bm{w}, \bm{z} | r, \bm{x}, \bm{y})].
\end{aligned}
\end{equation}

Taking the expectation under the variational posterior $q_{\theta}(\bm{w}, \bm{z}\mid r, \bm{x}, \bm{y})$:
\begin{equation}
\log p(r, \bm{x}, \bm{y}) = \mathbb{E}_{q_{\theta}} \left[\log \frac{p(\bm{w}, \bm{z}, r, \bm{x}, \bm{y})}{q_{\theta}(\bm{w}, \bm{z} \mid r, \bm{x}, \bm{y})}\right] + \mathrm{KL}[q_{\theta}(\bm{w}, \bm{z}) \| p(\bm{w}, \bm{z}\mid r, \bm{x}, \bm{y})].
\end{equation}

Since $\mathrm{KL}[q_{\theta}(\bm{w}, \bm{z}) \| p(\bm{w}, \bm{z}\mid r, \bm{x}, \bm{y})] \geq 0$, the ELBO is a lower bound:
\begin{equation}
\log p(r, \bm{x}, \bm{y}) \geq \mathcal{L}_{\text{ELBO}} := \mathbb{E}_{q_{\theta}} \left[\log \frac{p(\bm{w}, \bm{z}, r, \bm{x}, \bm{y})}{q_{\theta}(\bm{w}, \bm{z})}\right].
\end{equation}

Decomposing the joint distribution using the factorization $p(\bm{w}, \bm{z}, r, \bm{x}, \bm{y}) = p(\bm{w})p(\bm{z})p(r, \bm{x}, \bm{y} \mid \bm{w}, \bm{z})$ and the variational factorization $q_{\theta}(\bm{w}, \bm{z} \mid r, \bm{x}, \bm{y}) = q_{\theta_1}(\bm{w} \mid \bm{x})q_{\theta_2}(\bm{z} \mid \bm{x}, \bm{y})$:
\begin{equation}
\begin{aligned}
\mathcal{L}_{\text{ELBO}} &= \int_{\bm{w},\bm{z}} q_{\theta}(\bm{w}, \bm{z} | r, \bm{x}, \bm{y})\log \frac{p(\bm{w}, \bm{z}, r, \bm{x}, \bm{y})}{q_{\theta}(\bm{w}, \bm{z} | r, \bm{x}, \bm{y})}\mathrm{d}\bm{w}\mathrm{d}\bm{z} \nonumber \\
&= \mathbb{E}_{q_{\theta}(\bm{w}, \bm{z} | r, \bm{x}, \bm{y})}[\log \frac{p(\bm{w}, \bm{z}, r, \bm{x}, \bm{y})}{q_{\theta}(\bm{w}, \bm{z} | r, \bm{x}, \bm{y})}] \\
&= \mathbb{E}_{q_{\theta}(\bm{w}, \bm{z} | r, \bm{x}, \bm{y})} \left[ \log \frac{p(\bm{w})p(\bm{z})p(r, \bm{x}, \bm{y} \mid \bm{w}, \bm{z})}{q_{\theta_1}(\bm{w} \mid \bm{x})q_{\theta_2}(\bm{z} \mid \bm{x}, \bm{y})} \right] \\
&= \mathbb{E}_{q_{\theta}(\bm{w}, \bm{z} | r, \bm{x}, \bm{y})} [\log p(r, \bm{x}, \bm{y} \mid \bm{w}, \bm{z})] \nonumber 
+ \mathbb{E}_{q_{\theta_1}(\bm{w} \mid \bm{x})q_{\theta_2}(\bm{z} \mid \bm{x}, \bm{y})} \left[ \log \frac{p(\bm{w})p(\bm{z})}{q_{\theta_1}(\bm{w} \mid \bm{x})q_{\theta_2}(\bm{z} \mid \bm{x}, \bm{y})} \right]. \\
&= \mathbb{E}_{q_{\theta}(\bm{w}, \bm{z} | r, \bm{x}, \bm{y})} [\log p(r, \bm{x}, \bm{y} \mid \bm{w}, \bm{z})] 
- \mathbb{E}_{q_{\theta_1}(\bm{w} \mid \bm{x})} \left[\log \frac{q_{\theta_1}(\bm{w} \mid \bm{x})}{p(\bm{w})}\right] 
- \mathbb{E}_{q_{\theta_2}(\bm{z} \mid \bm{x}, \bm{y})} \left[\log \frac{q_{\theta_2}(\bm{z} \mid \bm{x}, \bm{y})}{p(\bm{z})}\right] \\
&= \mathbb{E}_{q_{\theta}(\bm{w}, \bm{z} | r, \bm{x}, \bm{y})} [\log p(r, \bm{x}, \bm{y} \mid \bm{w}, \bm{z})] 
- \mathrm{KL}[q_{\theta_1}(\bm{w} \mid \bm{x}) \| p(\bm{w})] 
- \mathrm{KL}[q_{\theta_2}(\bm{z} \mid \bm{x}, \bm{y}) \| p(\bm{z})].
\end{aligned}
\end{equation}
Since $\bm{x}$ and $\bm{y}$ are observed variables and are conditionally independent of $\bm{w}$ and $\bm{z}$, the conditional probability in the above equation can be simplified to $p(r \mid \bm{w}, \bm{z})$:
\begin{equation}
\mathcal{L}_{\text{ELBO}} = \mathbb{E}_{q_{\theta}(\bm{w}, \bm{z} | r, \bm{x}, \bm{y})} [\log p(r \mid \bm{w}, \bm{z})] 
- \mathrm{KL}[q_{\theta_1}(\bm{w} \mid \bm{x}) \| p(\bm{w})] 
- \mathrm{KL}[q_{\theta_2}(\bm{z} \mid \bm{x}, \bm{y}) \| p(\bm{z})].
\end{equation}
This completes the derivation of the ELBO for our causal reward model.

\section{Proof of Theorem~\ref{thm:pacbayes_vrm}}\label{app:pacbayes}
We give a proof by instantiating the PAC-Bayesian model averaging theorem of \citet{mcallester1999pacbayesian} with the latent-variable reward model.

For each preference triple $(\bm{x},\bm{y}^+,\bm{y}^-)$ and latent variables $(\bm{w},\bm{z}^+,\bm{z}^-)$, define the $0$-$1$ loss
$\ell(\bm{x},\bm{y}^+,\bm{y}^-)\in[0,1]$ as in Eq.~(\ref{eq:pref_01_loss}).

We use the factorized prior $p(\bm w, \bm{z}^+, \bm{z}^-)=p(\bm{w})p(\bm{z}^+)p(\bm{z}^-)$.
Given training data, our inference networks define the posterior
\begin{equation}
Q(\bm w, \bm{z}^+, \bm{z}^-) := q_{\phi_1}(\bm{w}\mid \bm{x})\,q_{\phi_2}(\bm{z}^+\mid \bm{x},\bm{y}^+)\,q_{\phi_2}(\bm{z}^-\mid \bm{x},\bm{y}^-).
\end{equation}

By the variational factorization and the factorized prior,
\begin{equation}
\begin{aligned}
&\mathrm{KL}\!\left(Q(\bm w, \bm{z}^+, \bm{z}^-)\,\|\,P(\bm w, \bm{z}^+, \bm{z}^-)\right)
= \mathbb{E}_{\bm w, \bm{z}^+, \bm{z}^-}\!\left[\log \frac{q_{\phi_1}(\bm{w}\mid \bm{x})\,q_{\phi_2}(\bm{z}^+\mid \bm{x},\bm{y}^+)\,q_{\phi_2}(\bm{z}^-\mid \bm{x},\bm{y}^-)}
{p(\bm{w})p(\bm{z}^+)p(\bm{z}^-)}\right] \\
&= \mathrm{KL}\!\left(q_{\phi_1}(\bm{w}\mid \bm{x})\,\|\,p(\bm{w})\right)
+ \mathrm{KL}\!\left(q_{\phi_2}(\bm{z}\mid \bm{x},\bm{y}^+)\,\|\,p(\bm{z})\right)
+ \mathrm{KL}\!\left(q_{\phi_2}(\bm{z}\mid \bm{x},\bm{y}^-)\,\|\,p(\bm{z})\right),
\end{aligned}
\end{equation}
which is exactly the requested separation into the $\bm{w}$-divergence and the $\bm{z}$-divergence.

Following \citet{mcallester1999pacbayesian}, we start with the following lemma:

\begin{lemma}
For $\ln\frac{1}{\delta} \leq 2N$, with probability at least $1-\delta$ over the selection of a sample $\mathcal{S}$ of size $N$, the following holds for all distributions $Q$ satisfying $\mathrm{KL}(Q\|P) \leq 2N$:
\begin{equation}
\mathcal{R}(Q) \leq \widehat{\mathcal{R}}(Q, \mathcal{S}) + \sum_{i} Q_i \sqrt{\frac{\ln\frac{Q_i}{P_i} + \ln\frac{1}{\delta} + \frac{5}{2}\ln N + 8}{2N-1}}
\end{equation}
\end{lemma}

In our case, the distribution $Q$ corresponds to the posterior over latent variables $Q(\bm{w}, \bm{z}^+, \bm{z}^-) = q_{\phi_1}(\bm{w}\mid \bm{x})\,q_{\phi_2}(\bm{z}^+\mid \bm{x},\bm{y}^+)\,q_{\phi_2}(\bm{z}^-\mid \bm{x},\bm{y}^-)$
and the prior is:$P(\bm{w}, \bm{z}^+, \bm{z}^-) = p(\bm{w})p(\bm{z}^+)p(\bm{z}^-)$

To obtain our main theorem from the lemma, we apply Jensen's inequality. Since the square root function is concave, we have:
\begin{equation}
\begin{aligned}
&\sum_{i} Q_i \sqrt{\frac{\ln\frac{Q_i}{P_i} + \ln\frac{1}{\delta} + \frac{5}{2}\ln N + 8}{2N-1}} \\
&\leq \sqrt{\frac{\sum_{i} Q_i \ln\frac{Q_i}{P_i} + \ln\frac{1}{\delta} + \frac{5}{2}\ln N + 8}{2N-1}} 
\end{aligned}
\end{equation}

Using the KL decomposition:
\begin{equation}
\begin{aligned}
\mathrm{KL}(Q\|P) &= \mathrm{KL}\!\left(q_{\phi_1}(\bm{w}\mid \bm{x})\,\|\,p(\bm{w})\right) \\
&\quad + \mathrm{KL}\!\left(q_{\phi_2}(\bm{z}\mid \bm{x},\bm{y}^+)\,\|\,p(\bm{z})\right) \\
&\quad + \mathrm{KL}\!\left(q_{\phi_2}(\bm{z}\mid \bm{x},\bm{y}^-)\,\|\,p(\bm{z})\right)
\end{aligned}
\end{equation}

Therefore, with probability at least $1-\delta$ over the draw of the training sample $\mathcal{S}$:
\begin{equation}
\begin{aligned}
\mathcal{R} &\leq \widehat{\mathcal{R}} + \sqrt{\frac{D(Q\| P) + \ln\frac{1}{\delta} + \frac{5}{2}\ln N + 8}{2N-1}} \\
&= \widehat{\mathcal{R}} + \Bigg(\frac{1}{2N-1}\Big(
\sum_i \\
&\quad\Big[\mathrm{KL}\!\left(q_{\phi_1}(\bm{w}\mid \bm{x}_i)\,\|\,p(\bm{w})\right) \\
&\quad + \mathrm{KL}\!\left(q_{\phi_2}(\bm{z}\mid \bm{x}_i,\bm{y}_i^+)\,\|\,p(\bm{z})\right) \\
&\quad + \mathrm{KL}\!\left(q_{\phi_2}(\bm{z}\mid \bm{x}_i,\bm{y}_i^-)\,\|\,p(\bm{z})\right)\Big] \\
&\quad + \ln\frac{1}{\delta} + \frac{5}{2}\ln N + 8\Big)\Bigg)^{1/2}
\end{aligned}
\end{equation}

This completes the proof of Theorem~\ref{thm:pacbayes_vrm}.

\paragraph{Remark.} Note that if $\ln\frac{1}{\delta} > 2N$ or $\mathrm{KL}(Q\|P) > 2N$, then the bound follows trivially from the fact that $\mathcal{R}(Q) < 1$. 
We focus on the meaningful range of the theorem and thus ignore these cases.
When both $\ln\frac{1}{\delta} \leq 2N$ and $\mathrm{KL}(Q\|P) \leq 2N$, the above proof applies.

\end{document}